\documentclass[letterpaper, 10pt, conference]{./ieeeconf}
\usepackage{times}
\usepackage[pdftex]{graphicx}

\usepackage{amsmath,amssymb,amsopn,amstext,amsfonts, amsthm}
\usepackage{cancel}
\usepackage[space]{cite}
\usepackage{pdfsync}
\usepackage{balance}
\usepackage{color}
\usepackage{mathtools}
\usepackage[ruled,norelsize, linesnumbered]{algorithm2e}
\usepackage{bm}
\usepackage{diagbox}
\usepackage{float}
\usepackage[outdir=./]{epstopdf}
\usepackage{url}
\usepackage{multirow}
\usepackage{makecell}
\usepackage{adjustbox}
\usepackage{tabularx, booktabs}
\let\labelindent\relax
\usepackage{enumitem}
\usepackage[font=small,skip=2pt]{caption}
\usepackage[linkcolor=black,citecolor=black,urlcolor=black,colorlinks=true]{hyperref}
\usepackage{subcaption}
\usepackage{bookmark}

\theoremstyle{remark}

\newcommand{\func}{\mathtt}

\SetKw{Continue}{continue}
\makeatletter
\newcommand{\removelatexerror}{\let\@latex@error\@gobble}
\makeatother

\graphicspath{{./}}
\DeclareGraphicsExtensions{.pdf,.png,.jpg,.eps}
\IEEEoverridecommandlockouts
\title{\LARGE \bf Efficient Uncertainty-aware Decision-making for Automated Driving Using Guided Branching}
\author{Lu Zhang$^\ast$, Wenchao Ding$^\ast$, Jing Chen, and Shaojie Shen%
\thanks{$^\ast$These authors contributed equally to this work. All authors are with the Department of Electronic and Computer Engineering, Hong Kong University of Science and Technology, Hong Kong, China.
{\tt\small lzhangbz@ust.hk, wdingae@ust.hk, jchenbr@ust.hk, eeshaojie@ust.hk}
This work was supported in part by the HKUST-DJI Joint Innovation Laboratory and in part by the Hong Kong Ph.D. Fellowship Scheme.}%
}

\begin{document}

\maketitle
\thispagestyle{empty}
\pagestyle{empty}

\begin{abstract}
    Decision-making in dense traffic scenarios is challenging for automated vehicles (AVs) due to potentially stochastic behaviors of other traffic participants and perception uncertainties (e.g., tracking noise and prediction errors, etc.). Although the \textit{partially observable Markov decision process} (POMDP) provides a systematic way to incorporate these uncertainties, it quickly becomes computationally intractable when scaled to the real-world large-size problem.
	In this paper, we present an efficient uncertainty-aware decision-making (EUDM) framework, which generates long-term lateral and longitudinal behaviors in complex driving environments in real-time. The computation complexity is controlled to an appropriate level by two novel techniques, namely, the \textit{domain-specific closed-loop policy tree} (DCP-Tree) structure and \textit{conditional focused branching} (CFB) mechanism. The key idea is utilizing domain-specific expert knowledge to guide the branching in both action and intention space. The proposed framework is validated using both onboard sensing data captured by a real vehicle and an interactive multi-agent simulation platform. We also release the code of our framework to accommodate benchmarking.
\end{abstract}

\section{Introduction}\label{sec:introduction}

In recent years, work in the areas of multi-sensor perception, prediction, decision-making, trajectory planning and control has enabled automated driving in difficult environments with other traffic participants and obstacles. Reasoning about hidden intentions of other agents is the key capability for a safe and robust automated driving system. However, even given perfect perception, it is still challenging to make safe and efficient decisions due to uncertain and sometimes unpredictable intentions of other agents. The situation is even worse when considering other system uncertainties such as imperfect tracking results and prediction errors.

There has been extensive literature on decision-making under uncertainty. Partially observable Markov decision process (POMDP)~\cite{smallwood1973pomdp} provides a general and principled mathematical framework for planning in partially observable stochastic environments. However, due to the \textit{curse of dimensionality}, POMDP quickly becomes computationally intractable when the problem size scales~\cite{madani1999undecidability}.

To address the computation difficulties, online POMDP planning algorithms~\cite{ross2008online} interleave the planning and execution and only reason about in the neighborhood of the current belief. Online POMDP solvers such as POMCP~\cite{silver2010pomcp}, DESPOT~\cite{ye2017despot, cai2018hyp} and ABT~\cite{kurniawati2016online} have been proposed. The latest advances in the POMDP solvers are applied to many uncertainty-aware planning algorithms for automated vehicles~\cite{bai2014integrated, liu2015situation, bai2015intention, brechtel2014probabilistic, hubmann2018automated, hubmann2018belief}. However, despite that various simplifications and discretizations are applied, the efficiency of the existing methods is still inadequate for highly dynamic driving scenarios (see Sec.~\ref{sec:related_work} for a detailed study).

\begin{figure}[t]
	\vspace{+0.5cm}
	\centering
	\includegraphics[width=0.48\textwidth]{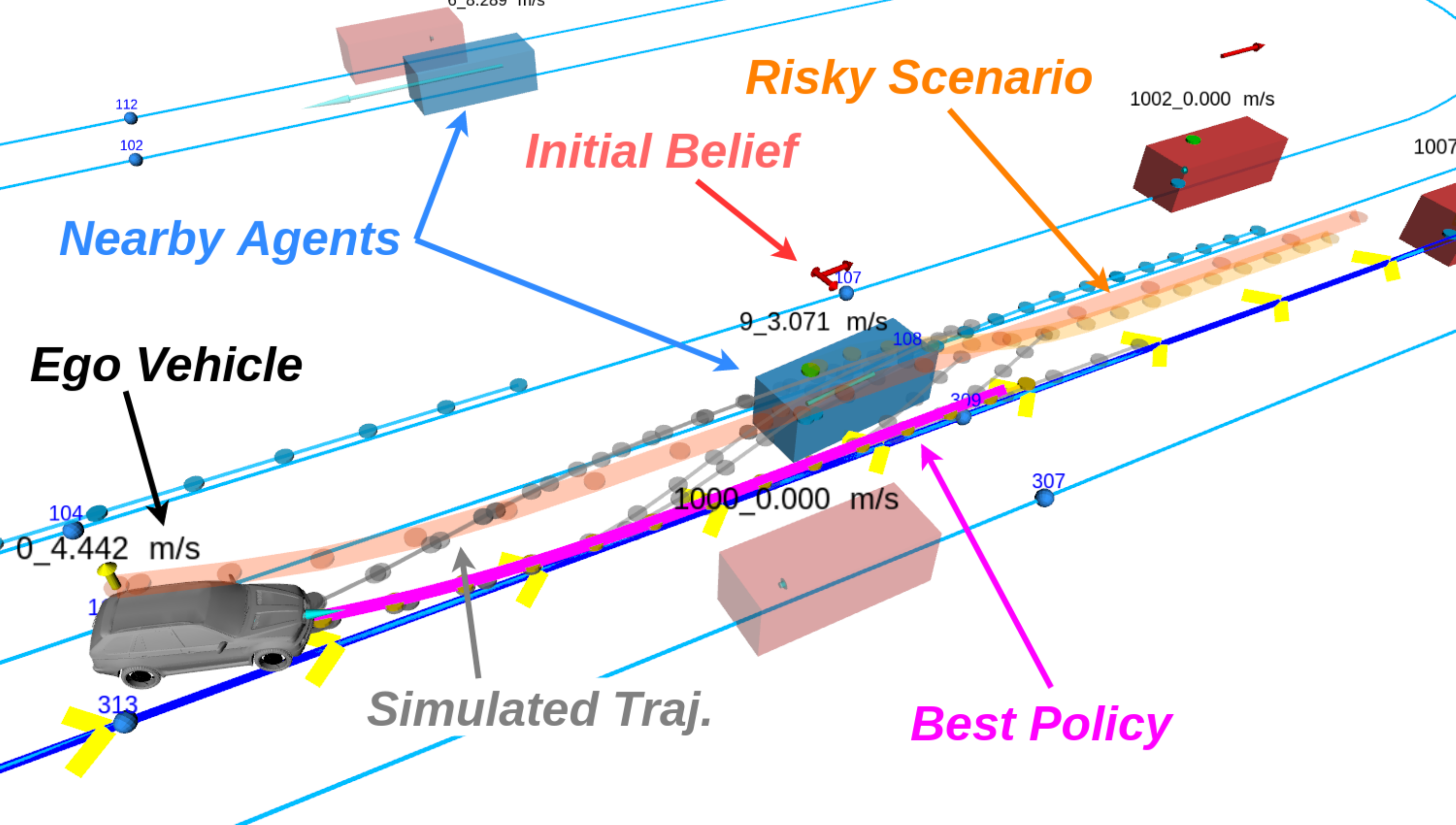}
	\caption{Illustration of the proposed decision-making framework. The initial belief for different intentions (i.e., LC, LK) is marked by the \textit{red} arrows on the agent vehicles with length representing the probability. The ego vehicle generates a set of sequential semantic-level policies according to the DCP-Tree. Each behavior sequence is simulated in closed-loop (marked by \textit{black} dots) considering nearby agents.
	The risky scenario (in \textit{orange}) in which the leading vehicle inserts to the ego target lane is identified by our CFB mechanism.}\label{fig:cover}
	\vspace{-0.5cm}
\end{figure}

It is essential to incorporate domain knowledge to efficiently make robust decisions. Multipolicy decision-making (MPDM)~\cite{cunningham2015mpdm, galceran2015mpdmchangept, galceran2017mpdmar} conducts deterministic closed-loop forward simulation of a finite discrete set of semantic-level policies (e.g., lane change (LC), lane keeping (LK), etc.) for the controlled (ego) vehicle and other agents, rather than performing the evaluation for every possible control input for every vehicle. However, the semantic behaviors for all the agents are assumed to be fixed in the whole planning horizon, which may not be true in long-term decision-making. Moreover, risk can be underestimated if initial behavior prediction is inaccurate~\cite{mehta2017fast, mehta2018backprop}, which may lead to unsafe decisions.

Our goal here is to control the computational complexity of the decision-making problem to enable real-time execution, while retaining sufficient flexibility and fidelity to preserve safety. In this paper, we present an efficient uncertainty-aware decision-making (EUDM) framework. First, EUDM uses a \textit{domain-specific closed-loop policy tree} (DCP-Tree) to construct a semantic-level action space. Each node in the policy tree is a finite-horizon semantic behavior of the ego vehicle. Each trace from the root node to the leaf node then represents a sequence of semantic actions of the ego vehicle. Each trace is evaluated in the form of closed-loop simulation similar to~\cite{cunningham2015mpdm} but the ego behavior is allowed to change in the planning horizon.

The DCP-Tree essentially determines a preliminary semantic-level action sequence for the ego vehicle, however, the behaviors (intentions) of other agent vehicles remain undetermined. Since the combinations of the intentions of agent vehicles explode exponentially, it is inefficient to naively sample all possible combinations of agent intentions.
To overcome this, EUDM uses the \textit{conditional focused branching} (CFB) mechanism to pick out the potentially risky scenarios using open-loop safety assessment conditioning on the ego action sequence. EUDM is highly \textit{parallelizable}, and can produce long-term (up to $8$ s) lateral and longitudinal fine-grained behavior plans in real-time ($20$ Hz).

The major contributions are summarized as follows:
\begin{itemize}
	\item An efficient uncertainty-aware decision-making framework for automated driving.
	\item A real-time and~\textit{open-source}\footnote{\url{https://github.com/HKUST-Aerial-Robotics/eudm_planner}} implementation of the proposed decision-making framework.
	\item Comprehensive experiments and comparisons are presented to validate the performance, using both onboard sensing data captured by a real vehicle and an interactive multi-agent simulation platform.
\end{itemize}

The remainder of this paper is organized as follows. The related work is reviewed in Section~\ref{sec:related_work}. An overview of the proposed decision-making framework is provided in Section~\ref{sec:system_overview}. The methodology and implementation are detailed in Section~\ref{sec:method} and Section~\ref{sec:implementation}, respectively.
Experimental results and benchmark analysis are elaborated in Section~\ref{sec:experimental_results}. Finally, this paper is concluded in Section~\ref{sec:conclusion}.

\section{Related Work}\label{sec:related_work}

There is extensive literature on decision-making for automated driving~\cite{schwarting2018planning, gonzalez2015review}. Many researchers tackle the planning problem in a decoupled manner, namely, ``predict and plan''~\cite{werling2012mp, mcnaughton2011motion, ziegler2014making, wei2014behavioral}. Specifically, prediction results are fixed in one planning cycle. Several drawbacks may exist. First, it is problematic to handle interaction among agents in this decoupled design. Second, given onboard sensing, imperfect tracking may result in prediction errors, which affects the safety of the decision. Third, even given perfect perception, the prediction uncertainty still scales dramatically w.r.t. the prediction horizon~\cite{ding2019predicting} due to partial observability.

POMDP is a powerful tool to handle various uncertainties in the driving task using a general probabilistic framework~\cite{kaelbling1998planning}. However, due to the \textit{curse of dimensionality}, POMDP quickly becomes computationally intractable when the problem size scales~\cite{madani1999undecidability}.

Leveraging the advance of online POMDP solvers~\cite{silver2010monte, ye2017despot, kurniawati2016online, cai2018hyp}, several methods are proposed to tackle the decision-making problem by simplifying the system model and restricting the problem scale. Bai~\textit{et al.}~\cite{bai2015intention} decouple the planning problem into pathfinding and velocity planning, and POMDP is only applied to the velocity planning. Hubmann~\textit{et al.} proposed POMDP-based decision-making methods for urban intersection~\cite{hubmann2018automated} and merging~\cite{hubmann2018belief} scenarios. However, human driving knowledge is not incorporated into the heuristic design. Meanwhile, the efficiency is still inadequate (less than 5 Hz) in fast-changing environments.

Cunninghan~\textit{et al.}~\cite{cunningham2015mpdm, galceran2017mpdmar} proposed the multipolicy decision-making (MPDM) framework, which approximates the POMDP process into the closed-loop simulation of predefined semantic-level driving policies (e.g., LC, LK, etc.) for all the agents. The incorporation of domain knowledge greatly accelerates the problem-solving. However, the ego behavior is fixed for the whole planning horizon, which may result in reactive decisions. Moreover, the hidden intentions (driving policy of other agents) are sampled according to initial behavioral prediction (initial belief) and will not be updated during the simulation. As a result, risky outcomes may not be reflected in policy evaluation due to inaccurate initial behavior prediction or insufficient intention samples~\cite{mehta2017fast}.

In this paper, we follow the idea of semantic-level closed-loop policies from MPDM. However, there are two major differences. First, the policy of the ego vehicle is allowed to change in the planning horizon according to the DCP-Tree, which makes it suitable for long-term decision-making. Second, focused branching is applied to pick out the risky scenarios, even given totally uncertain behavior prediction, which enhances the safety of the framework.

\begin{figure}[t]
	\centering
	\includegraphics[width=0.48\textwidth]{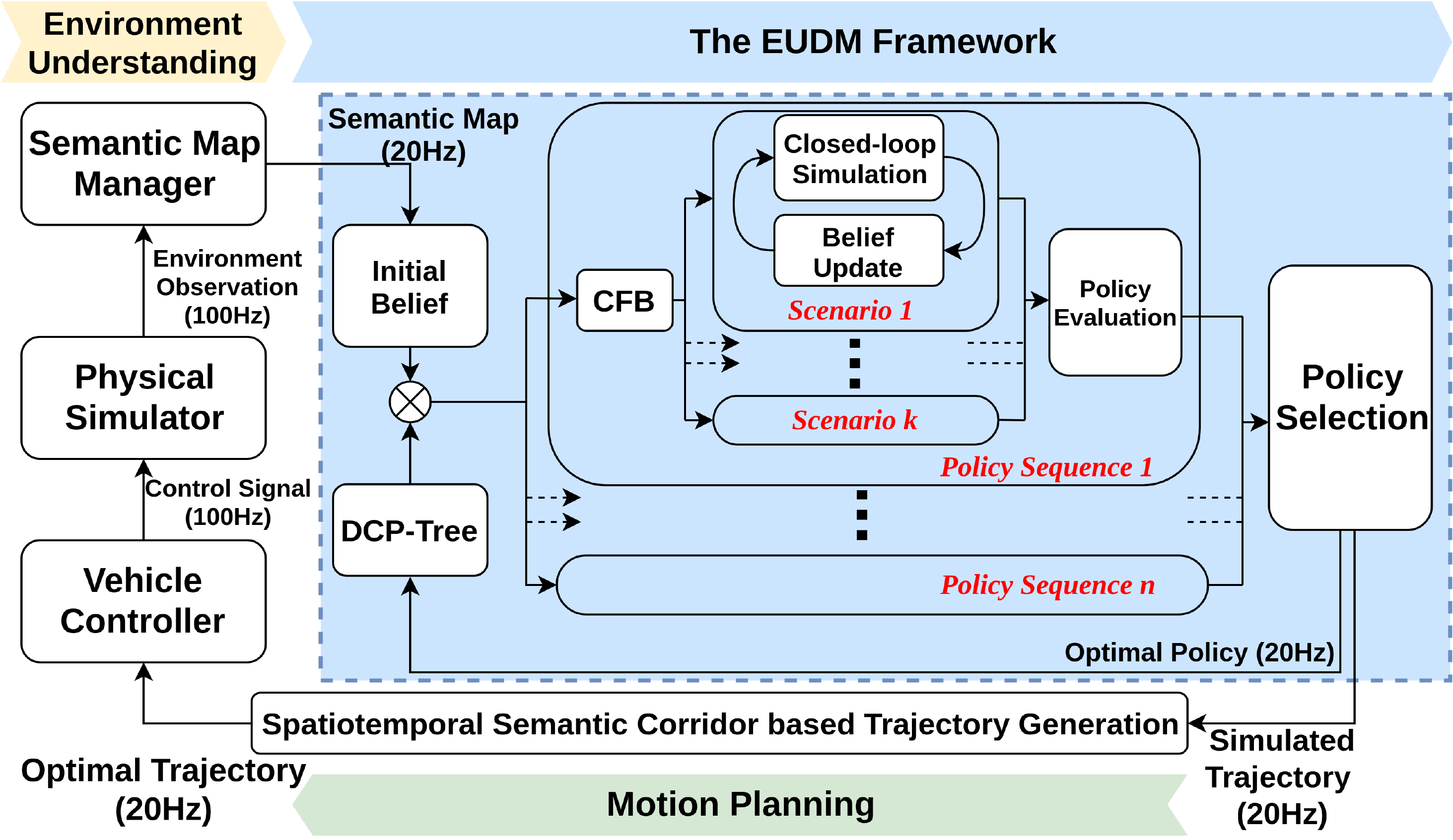}
	\caption{Illustration of the proposed decision-making framework (in the \textit{blue} box) and its relationship with other system components.}\label{fig:overview}
	\vspace{-0.8cm}
\end{figure}

\section{System Overview}\label{sec:system_overview}
An overview of our system is shown in Fig.~\ref{fig:overview}, which is similar to our previous work~\cite{ding2019safe}. The difference is that the focus of this paper is the decision-making part. Different from the methods which decouple the prediction and planning module~\cite{werling2012mp, mcnaughton2011motion, ziegler2014making, wei2014behavioral}, for our method, the intentions of agents are tracked and updated in the planning horizon.

In EUDM, DCP-Tree is used to guide the branching in the action domain and update the semantic-level policy tree based on the previous best policy. Each action sequence of the ego vehicle is then scheduled to a separate thread. For each ego action sequence, the CFB mechanism is applied to pick out risky hidden intentions of nearby vehicles and achieves guided branching in intention space.
The output of the CFB process is a set of scenarios containing different hidden intentions combinations of nearby vehicles.
Each scenario is then evaluated by the closed-loop simulation to account for interaction among agents in a sub-thread in parallel. All the scenarios are fed to the cost evaluation module and biased penalty is applied to risky branches. The output of the EUDM framework is the best policy which is represented by a series of discrete vehicle states ($0.4$ s resolution in the experiments) generated by the closed-loop forward simulation. The state sequence is fed to the motion planner to guide the trajectory generation process~\cite{ding2019safe}.

\section{Decision-making via Guided Branching}\label{sec:method}

\subsection{Preliminaries on POMDP}\label{sec:pomdp}

A POMDP can be defined as $\langle\mathcal{S}, \mathcal{A}, \mathcal{T}, \mathcal{R}, \mathcal{Z}, \mathcal{O}, \gamma\rangle$, which are the state space, action space, state-transition function, reward function, observation space, observation function, and discount factor, respectively. The state of the agent is~\textit{partially observable}, and is described as a belief $b$, which is a probability distribution over $\mathcal{S}$. The belief state can be updated given an action $a$ and an observation $z$ using Bayes' inference $b_t=\tau(b_{t-1}, a_{t-1}, z_t)$. The goal of the online POMDP planner is finding an optimal policy $\pi^*$ that maximizes the total expected discounted reward, given an initial belief state $b_0$ over the planning horizon $t_h$. We refer interested readers to \cite{kaelbling1998planning, silver2010pomcp} for more details.

The optimal policy is often pursued using a multi-step look-ahead search starting from the current belief $b_0$. A belief tree can be expanded using the \textit{belief update} function after taking actions and receiving observations during the search. 
However, the scale of the belief tree grows exponentially ($\mathcal{O}(|\mathcal{A}|^{h}|\mathcal{Z}|^{h})$) with respect to the tree depth $h$, which is computationally intractable given large action space $|\mathcal{A}|$ and observation space $|\mathcal{Z}|$. State-of-the-art online POMDP planners \cite{silver2010monte,ye2017despot,kurniawati2016online} use Monte-Carlo sampling to deal with the \textit{Curse of Dimensionality} and \textit{Curse of History}\cite{silver2010monte}.
Meanwhile, generic heuristic search such as branch-and-bound~\cite{ye2017despot} and reachability analysis~\cite{kurniawati2016online} can be used to accelerate the search. Note that the focus of this paper is to utilize domain-specific knowledge to achieve guided branching, which is also compatible with the generic heuristic search techniques.

\subsection{Domain-specific Closed-loop Policy Tree}
As pointed out by MPDM~\cite{galceran2015mpdmchangept}, for the decision-making problem, too much computation effort of POMDP is spent on exploring the space that is unlikely to be visited. The key feature of MPDM is using semantic-level policies instead of traditional ``state''-level actions (e.g., discretized accelerations or velocities). By using semantic-level policies, the exploration of the state space is guided by simple closed-loop controllers (i.e., domain knowledge).

\begin{figure}[t]
	\vspace{+0.5cm}
	\centering
	\includegraphics[width =0.48\textwidth]{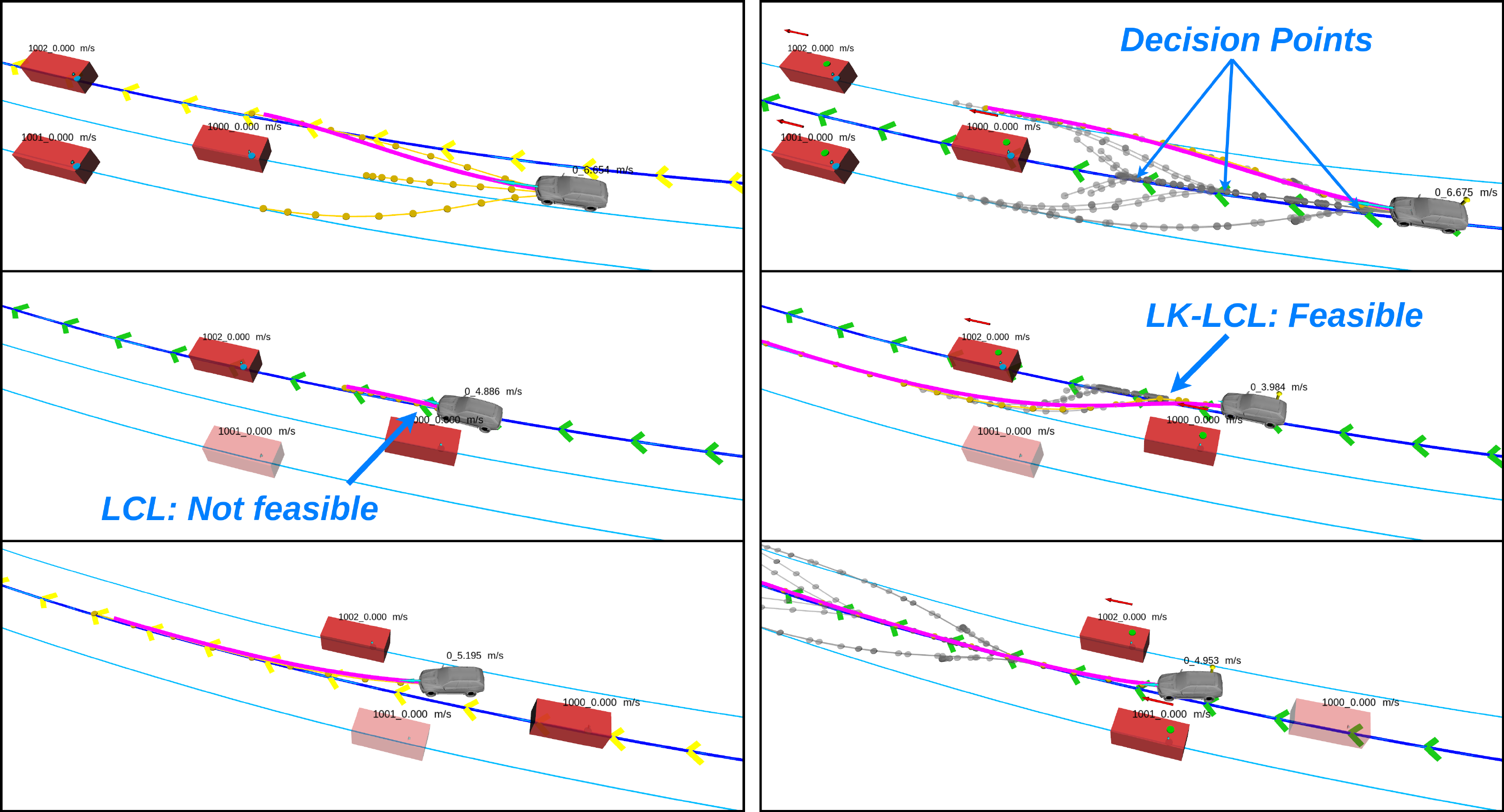}
	\caption{Comparison of MPDM (left) and EUDM (right). The ego vehicle (\textit{gray}) needs to conduct an overtaking maneuver. The simulated behaviors of MPDM are fixed in the planning horizon. The ego vehicle cannot make a lane-change-left (LCL) decision until it passes the blocking vehicle, so the generated plan is local and reactive.
	EUDM considers the change of behavior in different future stages, which results in a consistent and farsighted plan.}\label{fig:mpdm_eudm}
	\vspace{-0.4cm}
\end{figure}

Motivated by MPDM, we also use semantic-level policies, as one source of the domain knowledge. However, as elaborated in Sec.~\ref{sec:related_work}, one major limitation of MPDM is that the semantic-level policy of the ego vehicle is not allowed to change in the planning horizon. For example, MPDM may simulate LC and LK policies for the whole planning horizon (e.g., $8$ s). Typical patterns such as lane change in different stages (e.g., in $2$ s, $4$ s, $6$ s) are not included in its decision space. As a result, the decision of MPDM tends to be local and may not be suitable for long-term decision-making (see Fig.~\ref{fig:mpdm_eudm} for an example).

In this paper, DCP-Tree is utilized to generate future ego action sequences, which allows the semantic policy of the ego vehicle to change in the planning horizon.
The nodes of DCP-Tree are pre-defined semantic-level actions associated with a certain time duration. The directed edges of the tree represent the execution order in time. DCP-Tree origins from an \textit{ongoing} action $\hat{a}$, which is the executing semantic-level action from the best policy in the last planning cycle. Every time we enter a new planning episode, the DCP-Tree is rebuilt by setting $\hat{a}$ to the root node.

\begin{figure}[t]
	\centering
	\includegraphics[width=0.45\textwidth]{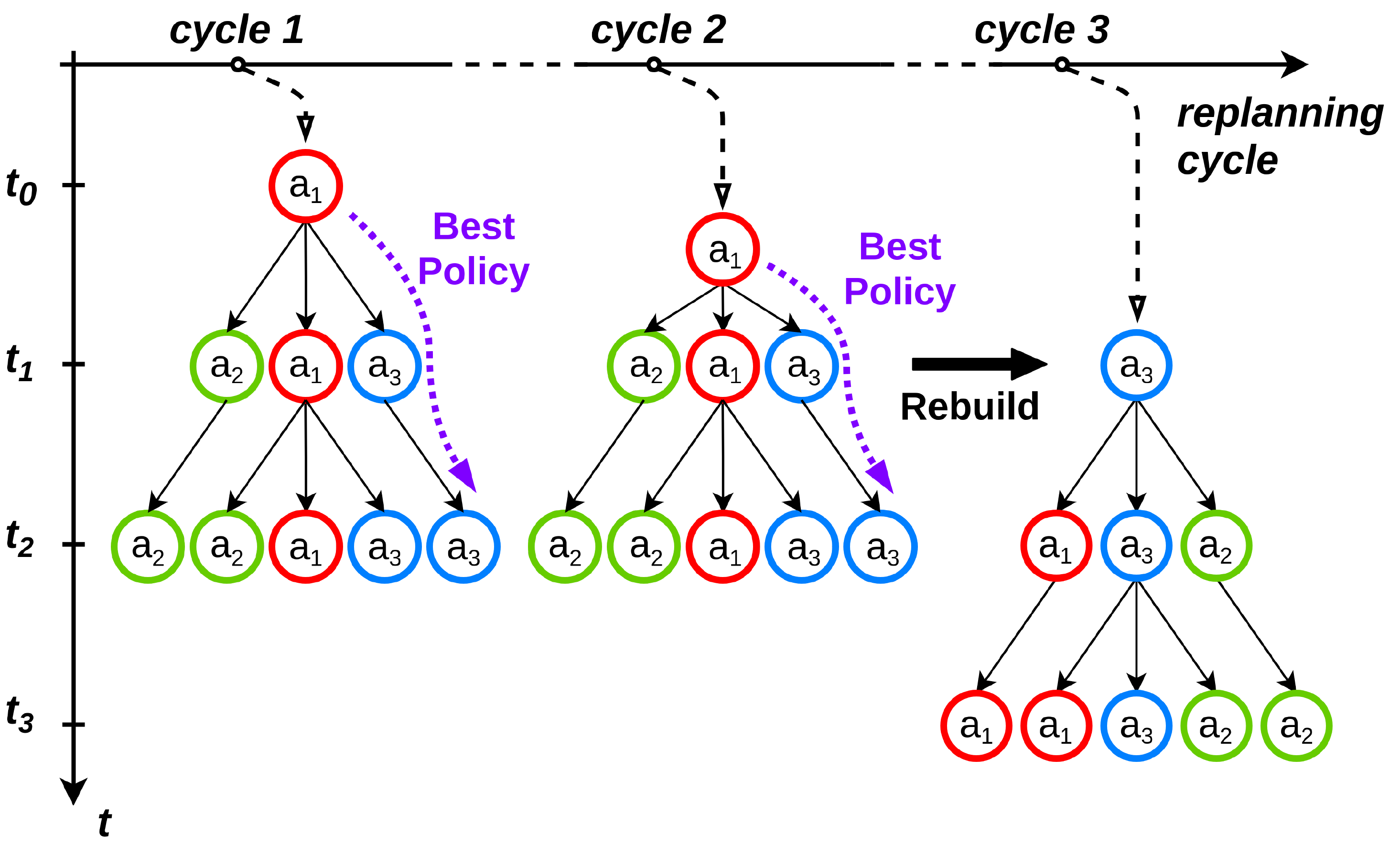}
	\caption{Illustration of the proposed DCP-Tree and the rebuilding process after its \textit{ongoing} action changes. Suppose there are three discrete semantic-level actions $\{a_1, a_2, a_3\}$ and the height of the tree is three. The \textit{ongoing} action for the left and middle tree is $a_1$, and the best policy is the dashed \textit{purple} trace. After executing $a_1$, the \textit{ongoing} action switches to $a_3$, and the DCP-Tree is updated to the right tree. Each trace only contains one change of action.}\label{fig:dcp_tree_mech}
	\vspace{-1.2cm}
\end{figure}

\begin{figure}[t]
	\removelatexerror
	\begin{minipage}{.48\textwidth}
		\begin{algorithm}[H]\label{algo:policy_selection}
			\caption{Process of EUDM}
			Inputs:
				Current states of ego and other vehicles $s$;
				\textit{Ongoing action} $\hat{a}$;
				Pre-defined semantic action set $\mathcal{A}$;
				Planning horizon $t_h$\;
			$\mathfrak{R}\leftarrow\emptyset$; $\slash\slash$ set of rewards for each policy\;
			$\Psi \leftarrow  \func{UpdateDCPTree(\mathcal{A},\hat{a})}$; $\slash\slash$ DCP-Tree $\Psi$\;
			$\hat{\Pi}\leftarrow\func{ExtractPolicySequences}(\Psi)$\;
			\ForEach{${\pi}\in\hat{\Pi}$}
			{
				$\Gamma^{{\pi}}\leftarrow\emptyset$; $\slash\slash$ set of simulated trajectories\;
				$\Omega \leftarrow\func{CFB}(s,{\pi})$; $\slash\slash$ set of critical scenarios\;
				\ForEach{$\omega \in \Omega$}{
					$\Gamma^{{\pi}}\leftarrow\Gamma^{{\pi}}\cup\func{SimulateForward}(\omega,{\pi},t_h)$\;
				}
				$\mathfrak{R}\leftarrow\mathfrak{R}\cup\func{EvaluatePolicy}({\pi},\Gamma^{{\pi}})$\;
			}
			${\pi}^*,\hat{a}\leftarrow\func{SelectPolicy}(\mathfrak{R})$\;
		\end{algorithm}
	\end{minipage}
	\vspace{-0.6cm}
\end{figure}

Since the ego policy is allowed to change in the planning horizon, the challenge is that the number of possible policy sequences scales exponentially w.r.t. the depth of the tree (i.e., the planning horizon).
To overcome this, DCP-Tree is expanded by a pre-defined strategy, which comes from the observation that, for human drivers, typically we do not frequently change the driving policy back and forth in a single decision cycle. For example, human drivers often evaluate whether it is feasible to conduct one policy change, e.g., switching from LK to LC in several seconds. This does not prevent human drivers from conducting complex maneuvers since consecutive decisions from different decision cycles can be combined. Motivated by this, from the~\textit{ongoing} action, each policy sequence will contain at most one change of action in one planning cycle, as shown in Fig.~\ref{fig:dcp_tree_mech}, while the back-and-forth behavior is achieved by replanning.

For instance, suppose the ongoing action is LK, the resulting policy sequences may include $($LK-LC-LC-LC$\ldots)$, $($LK-LK-LC-LC$\ldots)$ and $($LK-LK-LK-LC$\ldots)$, etc. Note that the size of leaf nodes in DCP-Tree is $\mathcal{O}[(|A|-1)(h-2) + |A|]$, $\forall h > 1$, which grows linearly with respect to the tree height $h$. It is also notable that MPDM is only one branch of our DCP-Tree and DCP-Tree includes multiple future decision points. Compared to MPDM, DCP-Tree has much larger decision space resulting in more flexible maneuvers as shown in Fig.~\ref{fig:mpdm_eudm}. Moreover, we also observe a significant improvement of decision consistency among consecutive planning cycles compared to MPDM.

\subsection{Conditional Focused Branching}
Essentially, DCP-Tree provides a guided branching mechanism in the action space.
The remaining problem is to determine the semantic-level intentions of the nearby vehicles, namely, the branching in the intention space. The challenge here is that the combination of intentions of nearby vehicles scales exponentially w.r.t. the number of agents. In the case of MPDM, the intention of the nearby vehicles is fixed for the whole planning horizon, and the initial intention is sampled according to a behavior prediction algorithm. The limitation of MPDM is that, with a limited number of samples, influential risky outcomes may not be rolled out, especially when the initial intention prediction is inaccurate~\cite{mehta2017fast}.

To overcome this, we propose the CFB mechanism. The goal here is to find the intentions of nearby vehicles which potentially lead to risky outcomes with as few branches as possible.
The term ``conditional'' means conditioning on the ego policy sequence. The motivation comes from the observation that the attention of the human driver for nearby vehicles is biased differently when intending to conduct different maneuvers. For example, a driver will pay much more attention to the situation on the left lane rather than the right one when he intends to make an LCL. As a result, by conditioning on the ego policy sequence, we can pick out a set of relevant vehicles w.r.t. the ego future actions. The selection process is currently based on rule-based expert knowledge as detailed in Sec.~\ref{sec:implementation}. We point out that learning-based attention mechanisms can also be incorporated and we leave it as an important future work.

By conditioning on the ego policy sequence, we obtain a subset of vehicles that need to be further examined. Instead of enumerating all the possible intentions for this subset of vehicles, we introduce a preliminary safety check to pick out the vehicles which we should pay special attention to. The preliminary safety assessment is conducted using open-loop forward simulation based on~\textit{multiple hypotheses}. For example, for the vehicle whose intention is uncertain, we anticipate what the situation will be if the vehicle is LC or LK, respectively. The anticipation is carried out using open-loop forward simulation under the intention hypothesis. The idea of using open-loop simulation is that by ignoring the interactions among agents, we check how the serious the situation will be if surrounding agents are completely uncooperative and does not react to the other agents. For the vehicles which do not pass the preliminary safety assessment, different scenarios are further examined in closed-loop forward simulation. And for the vehicles which pass the assessment, we use maximum a posteriori (MAP) from initial belief. As a result, the branching in intention space is guided to potentially risky scenarios. In practice, we find that the preliminary safety check can identify many dangerous cases despite its simple design.

The flow of EUDM is described in Algo.\ref{algo:policy_selection}. Evaluation for each policy sequence can be carried out in parallel (Line 5 to 11). Each critical scenario selected by CFB is examined by closed-loop forward simulation (Line 8 to 10) in parallel. Each policy is evaluated (Line 11) using the reward function detailed in Sec.~\ref{sec:implementation} and the best policy is elected (Line 13).

\section{Implementation Details}\label{sec:implementation}

\subsection{Semantic-level Actions}
We consider both lateral and longitudinal actions to ensure the diversity of the driving policy. Similar to~\cite{galceran2017mpdmar}, we define the lateral actions as~\{\textit{LK, LCL, LCR}\}. For longitudinal action, we use~\{\textit{accelerate, maintain speed, decelerate}\}. Note that these longitudinal actions are not discretized control signals such as acceleration commands in~\cite{hubmann2018automated, bai2015intention} but continuous desired velocity applied to the forward simulation model. Each semantic-level action is assigned with time duration of $2$ s, while the closed-loop simulation is carried out with $0.4$ s resolution to preserve the simulation fidelity. Note that the duration of the ongoing action is deducted by the replanning resolution ($0.05$ s) for each planning cycle. The depth of the DCP-Tree is set as $4$, thereby we obtain a planning horizon up to $8$ s.

\subsection{Forward Simulation}
The goal of the closed-loop simulation is to push the state of the multi-agent system forward while considering the potential interaction. The simulation model should achieve a good balance between simulation fidelity and inference efficiency. We adopt the \textit{intelligent driving model}~\cite{treiber2000congested} and~\textit{pure pursuit controller}~\cite{coulter1992implementation} as the longitudinal and lateral simulation models, respectively. Control noises are injected to reflect the stochastic property of driving behaviors.

\subsection{Belief Update}
The hidden intentions considered in this work include lateral behaviors, such as~\{\textit{LK, LCL, LCR}\}. The belief over these intentions of agent vehicles are updated during the forward simulation as shown in Fig.~\ref{fig:overview}.
In this work, we adopt a rule-based lightweight belief tracking module that takes a set of features and metrics including velocity difference, distance w.r.t. the leading and following agents on the current and neighboring lanes, responsibility-sensitive safety (RSS)~\cite{shalev2017formal} and lane-changing model~\cite{kesting2007general} as input\footnote{Detailed implementation can be found in our open-source code.}.
The belief tracker generates a probability distribution over the intentions (i.e., LK, LCL, LCR). The probability serves as the importance weight during policy evaluation. In the experiments, we find the rule-based belief tracker works well despite its simple structure. Currently, we are exploring using learning-based belief trackers for intention tracking~\cite{ding2019predicting} which will be incorporated into the EUDM framework.

\begin{figure}[ht]
	\centering
	\includegraphics[width =0.45\textwidth]{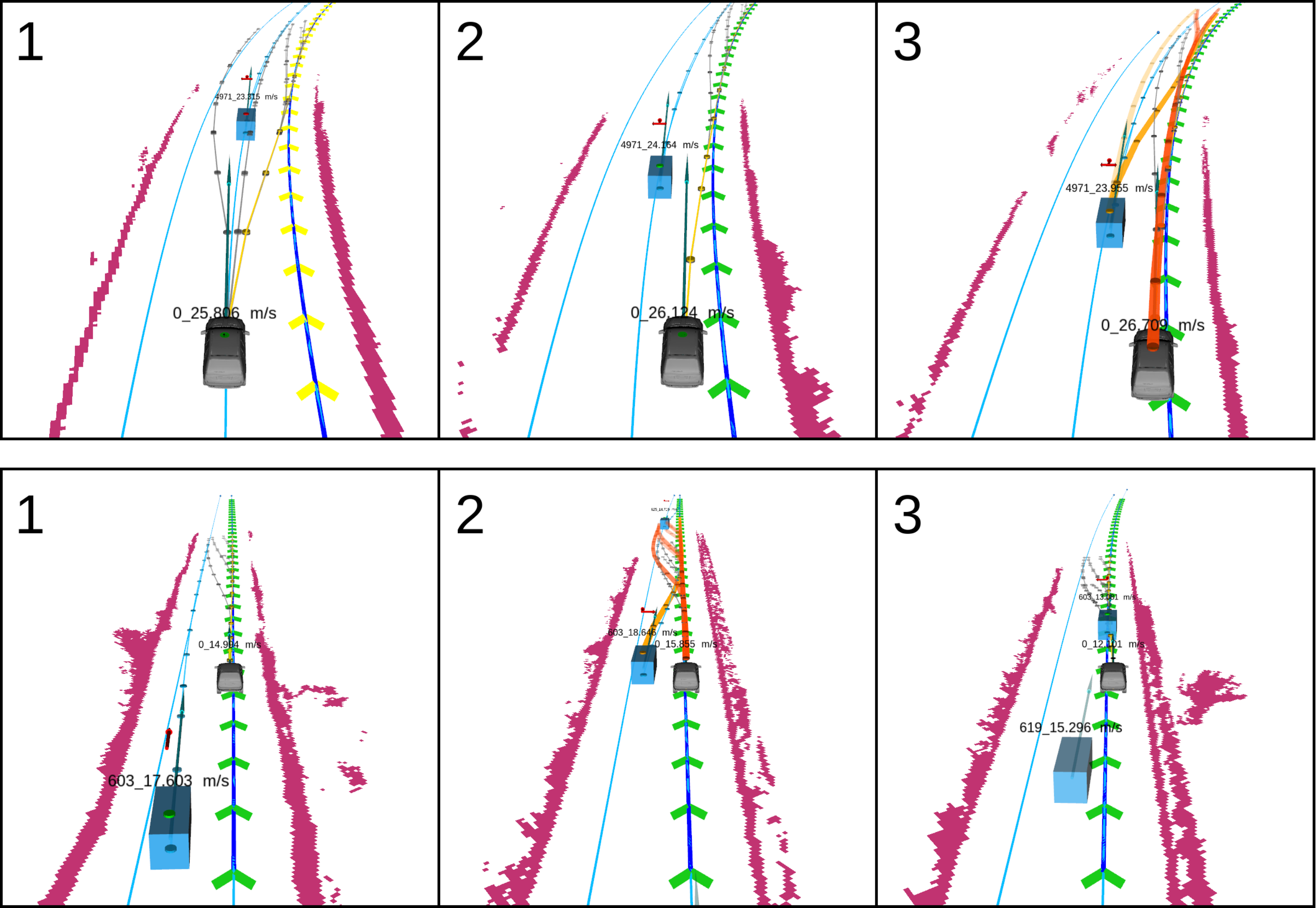}
	\caption{Open-loop test using onboard sensing data. Up: The ego vehicle is approaching another vehicle and making a LC decision (1 and 2). After the ego vehicle finishing LCR, EUDM detects several risky scenarios due to the uncertain intention of the front vehicle. Bottom: Another vehicle overtakes the ego vehicle and merges into our lane aggressively (1 and 2). EUDM captures the risky situation and takes a proper deceleration policy (3).}\label{fig:real_data}
	\vspace{-0.3cm}
\end{figure}

\subsection{CFB Mechanism}
The first step of CFB is the key vehicle selection. For the current lane and neighboring lanes, we search forward and backward along the lanes for a certain distance w.r.t. the current speed of ego vehicle and the vehicles in this range are marked as key vehicles. The second step is uncertain vehicle selection according to the initial belief. Specifically, we pick out the vehicles, whose probabilities for the three intentions are close to each other, as uncertain vehicles. Note that for the vehicles with confident prediction, we select the MAP intention and marginalize the intention probabilities using the MAP selection result. The third step is using the open-loop forward simulation for safety assessment. For the vehicles which fail the assessment, we enumerate all the possible combinations of their intentions. Each combination becomes a CFB-selected scenario and the probability of scenario is calculated. The fourth step is picking out top $k$ scenarios according to user-preference, and we further marginalize the probabilities among the top-$k$ scenarios. The marginal probabilities become the weights of CFB-selected scenarios during evaluation.

\begin{figure*}[t]
	\centering
	\hspace*{\fill}
	\begin{subfigure}[b]{0.99\textwidth}
		\centering
		\includegraphics[width =0.99\textwidth]{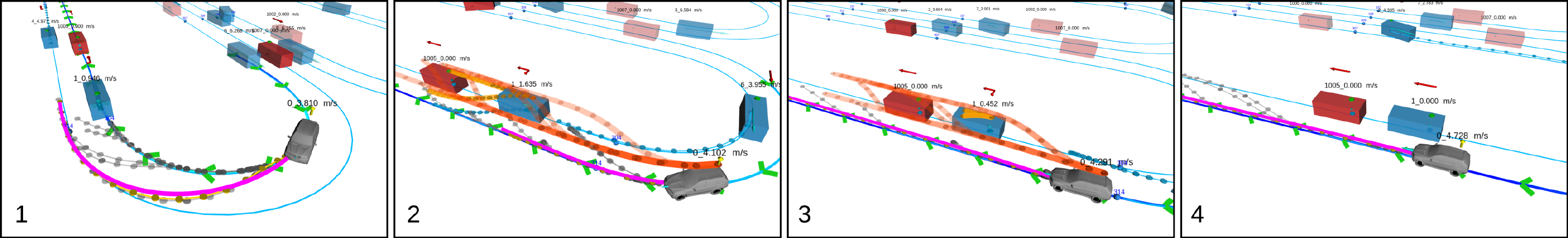}
		\caption{Overtake the leading vehicle. First, our vehicle generates an overtaking action and tries to LCL, since it confidently believes the leading vehicle will LK (1). Second, the leading vehicle accelerates as the ego vehicle approaching the zone, making it uncertain that whether the leading vehicle will insert before the ego vehicle. This risky scenario is identified by CFB and the ego vehicle plan to decelerate according to the closed-loop simulation (2). Third, the leading vehicle decelerates. Although the previous risky scenario still does not pass the preliminary safety assessment using open-loop simulation, the closed-loop simulation succeeds under the overtaking action by considering interactions. After evaluation, the ego vehicle decides to accelerate and pass.}\label{fig:overtake}
	\end{subfigure}
	\hspace*{\fill}
	\begin{subfigure}[b]{0.99\textwidth}
		\centering
		\includegraphics[width =0.99\textwidth]{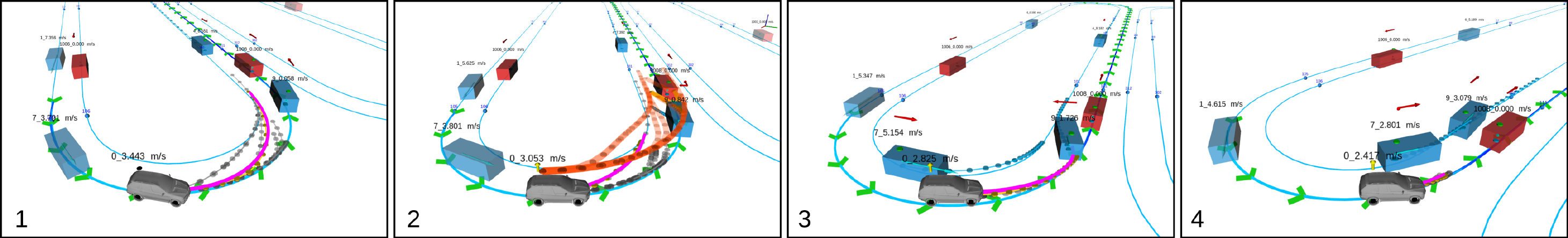}
		\caption{Give up overtaking and give way for other vehicles. First, our vehicle decides to LCL and overtake (1). Second, the leading vehicle accelerates, upon the assessment of potential risk, the ego vehicle decides to decelerate to follow the leading vehicle while conducting the LCL (2). Third, the leading vehicle conducts LCL and the following vehicle tries to overtake us (3). The ego vehicle then decides to give up LCL and yield (3 and 4).}\label{fig:yield}
	\end{subfigure}
	\caption{Illustration of different decision-making results in a conflict zone.}\label{fig:overtake_yield}
	\vspace{-0.6cm}
\end{figure*}

\subsection{Policy Evaluation}
The overall reward for a policy sequence is calculated by the weighted summation of the reward for each CFB-selected scenario. The reward function consists of a linear combination of multiple user-defined metrics including efficiency (measured by the difference between current velocity and desired velocity), safety (measured by the distance between our vehicle and surrounding vehicles) and consistency (measured by the difference between the last best policy and the policy to be evaluated).

\subsection{Trajectory Generation}\label{sec:mp}
The output of our behavior planner is a series of discrete states of the ego vehicle with 0.4 s resolution. The behavior plan is fed to the motion planner proposed in our previous work~\cite{ding2019safe}, which utilizes a spatio-temporal corridor structure to generate safe and dynamically feasible trajectories.

\section{Experimental Results}\label{sec:experimental_results}

\subsection{Simulation Platform and Environment}
The experiment is conducted in an interactive multi-agent simulation platform as introduced in Sect.\ref{sec:system_overview}. All agents can interact with each other without knowing the driving model of other vehicles. The forward simulation model used in the ego vehicle may differ from the actual model running on the agents.
The proposed decision-making method is implemented in C++11. All the experiments are conducted on a desktop computer equipped with an Intel i7-8700K CPU, and our proposed method can run stably at $20$ Hz.

\subsection{Qualitative Results}
To verify that our EUDM can generate flexible and consistent behaviors in highly interactive scenarios, we show several cases using both real-world data and simulation.

\subsubsection{Overtaking and Yielding in a Conflict Zone}
Conflict zones are common in urban traffic. As shown in Fig.\ref{fig:overtake_yield}, the ego vehicle has to pass through a conflict zone where there is a leading vehicle also trying to LC and pass through. Moreover, the initial belief of the leading vehicle is uncertain given the current observation.
As shown in Fig.\ref{fig:overtake_yield}, the EUDM framework can automatically select appropriate behaviors (i.e., overtaking or yielding) depending on the situation.

\begin{table}[ht]
	\vspace{-0.3cm}
	\centering
	\caption{Comparison of different decision-making approaches.\label{tab:dm_compare}}
	\begin{tabular}{@{}clcccc@{}}
	\toprule
	\multirow{2}{*}{\textbf{Map}} & \multirow{2}{*}{\textbf{{Method}}} &  \textbf{Safety} &  \textbf{Efficiency} & \multicolumn{2}{c}{\textbf{Comfort} ($\#/km$) }\\
																&                                    &                 & Ave. Vel. ($m/s$)   & UD & LCC \\
	\midrule
	\multirow{3}{*}{\textbf{ \makecell{Double\\Merge}}}  & MPDM & 0.048 & 4.9 & 1.85 & 4.63 \\
	                               & EDM  & 0.043 & \textbf{5.3} & 2.50 & 3.21 \\
	                               & EUDM & \textbf{0.025} & 4.9  & \textbf{0.38}  & \textbf{1.53} \\
  \midrule
	\multirow{3}{*}{\textbf{Ring}} & MPDM & 0.042 & 13.36 & 1.09 & 0.0 \\
	                               & EDM  & 0.030 & \textbf{14.37} & 1.41 & 0.0 \\
	                               & EUDM & \textbf{0.003} & 12.86 & \textbf{0.48} & 0.0 \\
  \bottomrule
  \end{tabular}
  \vspace{-0.3cm}
\end{table}

\subsubsection{Testing using Real-world Onboard Sensing Data}
In this case, our method is tested in an open-loop manner by using the data collected by a real automated vehicle. The goal is to verify whether the proposed method can capture risky scenarios and work under uncertain predictions and noisy perception. As shown in Fig.~\ref{fig:real_data}, EUDM can make appropriate decisions to overtake or decelerate depending on the situation.

\subsection{Quantitative Results}
We conduct a comprehensive quantitative comparison with the MPDM~\cite{galceran2017mpdmar}, which is one of the state-of-the-art decision-making methods for automated driving. We evaluate the two methods using two benchmark tracks, i.e., \textit{Double Merge} and \textit{Ring}. Detailed experiments can be found in the attached video, while the statistical results are shown in Table.~\ref{tab:dm_compare}.

\subsubsection{Metrics}
We introduce three major metrics to evaluate the performance of two methods, namely, safety, efficiency, and comfort. For safety, we count the fraction of frames that the distance between ego vehicle and other agents smaller than a threshold (i.e., safety distance). The efficiency is represented by the average velocity of the ego vehicle. The comfort is described by the number of \textit{uncomfortable deceleration} (UD) and the number of \textit{large curvature changing} (LCC) per kilometers. The threshold of UD and LCC is set to 1.6 $m/s^2$ and 0.12 $(s\cdot m)^{-1}$, respectively.

\subsubsection{Benchmarking}
We also conduct an ablative study by removing the CFB mechanism from EUDM, which results in the EDM method as shown in Table.~\ref{tab:dm_compare}. The goal is to verify whether the CFB can improve the robustness and safety of the framework.
As shown in Table.~\ref{tab:dm_compare}, EUDM makes safer decisions than EDM and MPDM according to the safety metric. The reason is that EUDM explicitly explores the risky scenarios and conducts biased branching. Note that the double merge map is a dense interactive scenario where unsafe situations are hard to be completely avoided due to aggressive behaviors of agent vehicles.
For efficiency, our method and MPDM perform similarly, while EDM has a higher average velocity. It is because EDM enlarges the action space compared to MPDM, and it may take over-aggressive risky actions (see Fig.\ref{fig:mpdm_eudm}). In terms of comfort, EUDM can generate much smoother behaviors than the other two baselines, since it takes conservative policy under risky scenarios beforehand and avoids hard brakes.

\section{Conclusion and Future Work}\label{sec:conclusion}
In this paper, we proposed the EUDM framework for automated driving in dense interactive scenarios, by introducing two novel techniques, namely, the DCP-Tree and CFB mechanism. The complete framework is open-sourced and comprehensive evaluations are conducted using both real-world data and simulation.
In the future, we will conduct closed-loop field test for the EUDM framework.

\end{document}